%% file: main.tex
\documentclass[10pt,twocolumn,letterpaper]{article}

\usepackage[pagenumbers]{cvpr} 
\usepackage{epigraph} 
\usepackage{xcolor}
\usepackage{multirow}
\usepackage{graphicx}
\usepackage{amsmath, amssymb}

\usepackage{soul}
\input{preamble}
\definecolor{cvprblue}{rgb}{0.21,0.49,0.74}
\usepackage[pagebackref,breaklinks,colorlinks,citecolor=cvprblue]{hyperref}

\def\paperID{30} 
\def\confName{3DV\xspace}
\def\confYear{2026\xspace}

\title{\ourmethod: Single-View 3D Stereo Reconstruction Aided by Mirror Reflections}

\author{Jing Wu \quad Zirui Wang \quad Iro Laina \quad Victor Adrian Prisacariu\\
University of Oxford\\
{\tt\small \{jing.wu, iro.laina, victor.prisacariu\}@eng.ox.ac.uk, ryan@robots.ox.ac.uk}
}

\begin{document}



\input{sec/00_teaser}
\input{sec/0_abstract}    
\input{sec/1_intro}
\input{sec/2_relatedworks}
\input{sec/3_method}

\input{sec/4_dataset}

\input{sec/4_experiments}
\input{sec/5_conclusion}
{
    \small
    \bibliographystyle{ieeenat_fullname}
    \bibliography{main}
}
\input{sec/X_suppl}

\end{document}

%% file: preamble.tex
\usepackage[dvipsnames]{xcolor}

\newcommand*{\ourmethod}{Reflect3r\@\xspace}
\newcommand*{\reconmodel}{DUSt3R\@\xspace}
\newcommand*{\synscenenum}{16\@\xspace}

\newcommand{\boldstart}[1]{\vspace{0pt}\noindent\textbf{#1}}

%% file: sec/00_teaser.tex
\twocolumn[{%
\renewcommand\twocolumn[1][]{#1}%
\maketitle
\begin{center}
    \centering
\includegraphics[width=.84\linewidth]{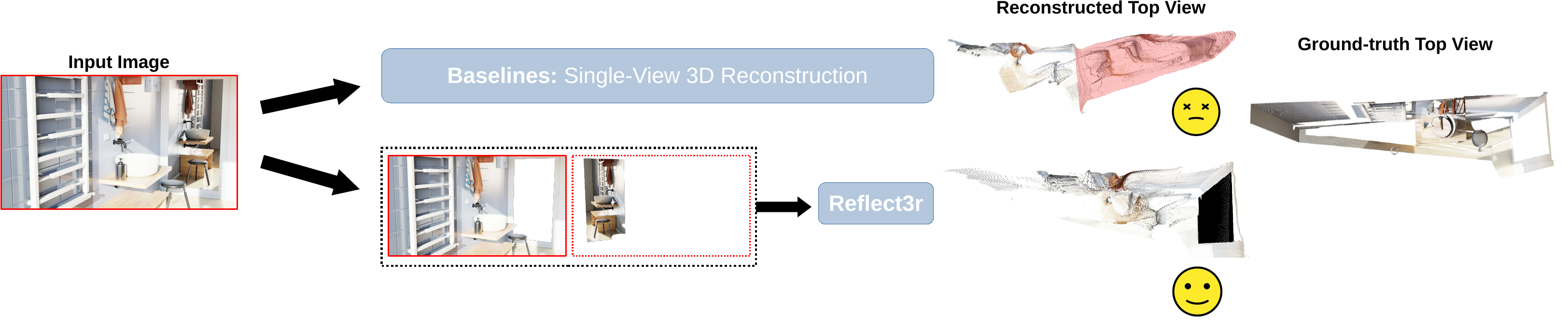}
    \captionof{figure}{\textbf{Single-view 3D reconstruction with mirror reflections.} Given an image containing a mirror, we aim to reconstruct the 3D geometry of the scene. 
    Existing methods cannot recognise the reflective cues and fail by predicting a false geometry for the mirror region, which is highlighted with \textcolor{purple}{light red} (Top). 
    We reinterpret the mirror reflection as a virtual view captured by a simulated camera, enabling a stereo formulation that leads to more accurate geometry reconstruction (Bottom).
    }
\label{fig:teaser}
\end{center}
}]

%% file: sec/0_abstract.tex
\begin{abstract}
\vspace{-0.6cm}
Mirror reflections are common in everyday environments and can provide stereo information within a single capture, as the real and reflected virtual views are visible simultaneously.
We exploit this property by treating the reflection as an auxiliary view and designing a transformation that constructs a physically valid virtual camera, allowing direct pixel-domain generation of the virtual view while adhering to the real-world imaging process.
This enables a multi-view stereo setup from a single image, simplifying the imaging process, making it compatible with powerful feed-forward reconstruction models for generalizable and robust 3D reconstruction.
To further exploit the geometric symmetry introduced by mirrors, we propose a symmetric-aware loss to refine pose estimation.
Our framework also naturally extends to dynamic scenes, where each frame contains a mirror reflection, enabling efficient per-frame geometry recovery.
For quantitative evaluation, we provide a fully customizable synthetic dataset of \synscenenum Blender scenes, each with ground-truth point clouds and camera poses.
Extensive experiments on real-world data and synthetic data are conducted to illustrate the effectiveness of our method. 
Code and Blender scenes are available in: \url{https://github.com/jingwu2121/reflect3r.git}.



\end{abstract}

%% file: sec/1_intro.tex
\section{Introduction}



\noindent Mirror reflections are ubiquitous in our daily environments. 
They allow multi-view cues, comprising the real scene and its mirrored counterparts, to be inferred from a single capture, naturally forming an epipolar geometry.
Unlike a classical multi-view setup, the real–virtual view pairs share the same intrinsic parameters, which simplifies calibration.
This property can significantly reduce hardware requirements, removing the need for cross-camera synchronization and shortening scanning time.


Effectively leveraging reflections for 3D reconstruction is non-trivial, as it involves a multi-view relationship derived from only a single image. 
Early mirror-based methods \cite{Stereowithmirrors, Mitsumoto1992mirror_recon} analytically reconstructed simple polyhedral shapes, but relied on highly controlled settings and thus do not generalize to real-world scenes with complex shapes, materials, and occlusions.
More recently, Kawahara et al. \cite{Kawahara2023waterref} explored using water reflections as indirect views by explicitly modeling light transport to recover scene geometry and appearance, reconstructing 3D point coordinates via standard stereo triangulation from correspondences in the direct–reflected pairs.
Fang et al. \cite{fang2021mirrored} explored human pose reconstruction by leveraging mirror reflections of human bodies, relying on 2D correspondence keypoints for initialisation.
However, both approaches are inflexible and degrade significantly when the reflected view has a large angular difference from the real view, as reliable correspondences become sparse. 
Moreover, they are tailored to highly specific scenarios, such as water reflections of outdoor buildings, simplified objects, or the human body, and lack the ability to generalize to diverse, real-world scenes.
Recent feed-forward reconstruction models \cite{dust3r_cvpr24, mast3r_arxiv24, wang2025vggt} achieve impressive accuracy, robustness, and generalisation across diverse settings, yet they lack awareness of reflected virtual views.
As a result, they fail to exploit reflections in single-view reconstruction and are often confused by them in this monocular setup.
To address this, we explicitly integrate reflection-derived virtual views into feed-forward reconstruction models, treating the problem as a stereo setup, formulating the problem as \textit{reflection-aided single-view 3D reconstruction}.
The goal of this stereo configuration is to reconstruct the full 3D geometry of the scene from a single RGB image capture containing a mirror, leveraging the inherent multi-view constraints introduced by mirror reflections. 
Our key insight is to \textbf{reinterpret reflected virtual views as auxiliary views that contribute complementary geometric and appearance cues}. 
This is non-trivial, as the reflected views need first to be transformed into physically valid virtual views and then combined with the real view to form a consistent multi-view setup.
Motivated by this, we propose \textbf{\ourmethod}, a reflection-aided single-image 3D reconstruction framework capable of operating at the scene level in unconstrained environments. 
As illustrated in \cref{fig:teaser}, \ourmethod substantially improves the coverage of 3D reconstruction.  
It recovers 3D geometry from a single image containing a mirror by treating the image as comprising two complementary views separated by the mirror surface. 

To achieve this, we design a multi-view setup configuration process that operates directly in the pixel domain to create a valid virtual view, simulating the real-world imaging process. 
In real-world scenarios, the real and reflected virtual views often exhibit large angular differences, making reconstruction challenging.
Recent large-scale reconstruction models \cite{dust3r_cvpr24} have demonstrated a strong ability to recover geometry even under sparse or wide-angle inputs.
Building on this, \ourmethod leverages \reconmodel, which directly predicts dense geometry from multi-view inputs.
However, \reconmodel alone fails in the mirror geometry prediction and often hallucinates depth within reflective regions.




To mitigate this, we provide \reconmodel with the pre-built multi-view setup, helping it better capture the spatial configuration.
This stereo formulation treats the reflection as an auxiliary camera and introduces a natural symmetry constraint.
To exploit this constraint, we design a symmetric-aware loss for post-optimization, refining the estimated camera poses by enforcing symmetry between the real and virtual poses with respect to the mirror plane.

Furthermore, our formulation naturally extends to dynamic scenes: given a video in which each frame contains a mirror reflection, \ourmethod can reconstruct per-frame geometry efficiently and at low capture cost.

Finally, existing 3D datasets with mirrors lack ground-truth for the virtual views.
To enable quantitative evaluation, we construct a fully customizable synthetic dataset in Blender \cite{blender}, consisting of \synscenenum manually created scenes with ground-truth point clouds and real/virtual camera poses.
%
The contributions of this work are as follows:
\begin{itemize}
    \item We reinterpret mirror reflections as auxiliary viewpoints, enabling a formulation of \textit{single-view stereo} for 3D reconstruction.
    \item We design a multi-view configuration process that works directly in the pixel domain to create a valid virtual view consistent with the physical imaging process.
    \item We exploit the inherent symmetry of mirrored scenes and propose a symmetric-aware loss to refine pose estimation.
    \item To facilitate quantitative evaluation, we contribute a synthetic dataset with \synscenenum ground-truth point clouds and poses, tailored for this task, fully customizable and easily extendable for future research.
\end{itemize}

Extensive experiments demonstrate that \ourmethod reconstructs a more complete 3D geometry from a single image cue compared to all baselines.

%% file: sec/2_relatedworks.tex
\section{Related Works}

\subsection{Computer Vision with Mirror}
\boldstart{Mirror Detection and Segmentation.}
Mirrors have long been treated as challenging elements in image understanding due to their ambiguous visual cues and deceptive appearances. 
In static image analysis, several works have focused on detecting and segmenting mirrors. 
DAM \cite{xing2023DAM}, Progressive Mirror Detection \cite{lin2020ProgressiveMirrorDetection}, and PDNet \cite{Mei2021PDNet} incorporate geometric and semantic priors to distinguish mirrors from ordinary surfaces. 
For video, Warren et al. \cite{Warren2024vid_mirror} and Lau et al. \cite{Lau_2024zoom} extend mirror segmentation into the temporal domain, leveraging motion consistency to improve accuracy.

\boldstart{Mirror Generation.}
Beyond detection, recent generative models have explored how to synthesize realistic reflections. 
MirrorVerse v1 and v2 \cite{dhiman2025mirrorverse, Ankit2025mirrorfusion} tackle this by modelling scene semantics to generate plausible mirror content during image generation.

\boldstart{Mirror Geometry and Depth Estimation.}
In 3D vision, mirrors are particularly problematic due to their impact on geometry and depth estimation. 
Classical approaches often misinterpret mirror reflections as real geometry, leading to incorrect scene understanding. 
To address this in laser scans, \cite{yang2008laser} proposed a sensor fusion technique for dealing with LiDAR sensor failures on mirror and glass surfaces.
\cite{Kshammer2015MIRRORIA} detects mirrors and corrects laser-scanned point clouds based on heuristics using known mirror dimensions.
Mirror3D \cite{mirror3d2021tan} corrected depth estimates in mirrored regions of both raw scanned and estimated depth by estimating the 3D mirror plane based on RGB input and surrounding depth context. 

\boldstart{Novel View Synthesis.}
In the context of novel view synthesis and 3D reconstruction, modeling reflective materials has gained attention. 
Works such as Ref-NeRF \cite{verbin2022refnerf}, NeRFReN \cite{Guo2022NeRFReN}, NeRSP \cite{nersp2024yufei}, and 3DGS-DR \cite{ye2024gsdr} extend radiance fields \cite{mildenhall2020nerf} and 3D Gaussian Splatting \cite{kerbl3Dgaussians} to better handle reflections in surfaces like glass and metal. 
Closer to our task, MirrorGaussian \cite{liu2024mirrorgaussian}, Mirror-NeRF \cite{zeng2023mirror-nerf}, and MS-NeRF \cite{Yin2025msnerf} explicitly target mirrors in 3D, aiming to reconstruct scenes with accurate mirror appearance and geometry. 
\cite{Kawahara2023waterref} explores the 3D reconstruction from water reflections. 

Most prior works treat reflections as noise, and attempts to leverage mirrors for reconstruction remain limited, despite the fact that reflections inherently provide a second viewpoint of the scene.
In contrast, our work incorporates mirror reflections by reinterpreting them as views captured by virtual cameras, reformulating the single-image cue as a multi-view setup, and enabling more complete 3D reconstruction even in unconstrained settings.

\subsection{Single-Image-to-3D Reconstruction} 
Single-image-to-3D methods can be broadly divided into two lines of work.

\boldstart{Generative.}
The first line recovers 3D geometry through generation by hallucinating multi-view images of a 3D model and reconstructing the scene from the synthesized dense views.
Early approaches, such as RealFusion \cite{melaskyriazi2023realfusion} and Make-It-3D \cite{tang2023makeit3d}, adopt a per-scene optimization strategy, distilling prior knowledge from 2D generative models into 3D representations via the Score Distillation Sampling (SDS) loss \cite{poole2022dreamfusion}.
While initially designed for object-level reconstruction, this paradigm has been extended to scene-level synthesis in works like Reconfusion \cite{wu2023reconfusion}.
More recent methods \cite{liu2023zero123, zheng2023free3D, kong2024eschernet, gao2024cat3d} bypass per-scene optimization by directly training diffusion models that, conditioned on an input view and target camera parameters, predict novel views which are then used for 3D reconstruction.

\boldstart{Reconstruction Aided by Reflection.}
The second line of work reconstructs 3D geometry directly from the given information without hallucination.
Early mirror-based methods \cite{Stereowithmirrors, Mitsumoto1992mirror_recon} reconstructed simple polyhedral shapes under highly controlled conditions, but it is unable to generalise to real-world scenes with complex geometry and materials.
Kawahara et al. \cite{Kawahara2023waterref} and Fang et al. \cite{fang2021mirrored} use the 2D correspondences between the real and the reflected water/mirror view to reconstruct a scene or a human body, respectively.  
However, these methods are inflexible and degrade significantly when the reflected view has a large angular difference from the real view, as reliable correspondences become sparse. 
Moreover, they are tailored to highly specific scenarios, such as water reflections of outdoor buildings, simplified objects, or the human body, and cannot generalize to diverse, real-world scenes.



In contrast, our work extends image-to-3D reconstruction by leveraging real visual cues, specifically reflections, treating them as auxiliary viewpoints to enhance single-image 3D reconstruction without relying on hallucinated geometry. 
Furthermore, we reformulate a valid multi-view setup by simulating the physical imaging process, ensuring compatibility with the backbone model and improving the quality of single-image reconstruction, thereby enabling generalization to real-world scenes.

%% file: sec/3_method.tex
\section{Method}

Given an input image $I \in \mathbb{R}^{H \times W \times 3}$ containing mirrors $\mathcal{F} = \{ F_1, F_2, \dots \}$, our goal is to reconstruct the 3D geometry of the scene as a point cloud. 
We achieve this by decomposing the single input image into a set of incomplete multi-views $\mathcal{I} = \{ I_{\text{real}}, I_{\text{vir}_1}, I_{\text{vir}_2}, \dots, I_{\text{vir}_N} \}$, where $I_{\text{real}}$ is the real view and $I_{\text{vir}_i}$ is the mirror reflection derived virtual view, and $N$ is the number of the virtual views. 
Each virtual view provides complementary information about the scene.


Specifically, for each real-virtual view pair $(I_{\text{real}}, I_{\text{vir}_i})$, there exists a mirror plane $\textbf{M}_i = (\mathbf{n}_i, \mathbf{p}_i)$, where $\mathbf{n}_i \in \mathbb{R}^{3\times1}$ is the normal vector and $\mathbf{p}_i \in \mathbb{R}^{3\times1}$ is a point on the plane, such that the corresponding real and virtual camera poses $\mathbf{C}_{\text{real}} = [\textbf{R}_{\text{real}} | \textbf{t}_{\text{real}}] \in \mathbb{R}^{4\times4}$ and $\mathbf{C}_{\text{vir}_i} = [\textbf{R}_{\text{vir}_i} | \textbf{t}_{\text{vir}_i}] \in \mathbb{R}^{4\times4}$ are related by the corresponding reflection transformation matrix $\mathbf{T}_{\text{reflect}_i}$:
\begin{align}
    \mathbf{C}_{\text{vir}_i} = \mathbf{T}_{\text{reflect}_i} \mathbf{C}_{\text{real}}.
\end{align}
This geometric constraint ensures that the real and virtual cameras share a symmetric configuration relative to the mirror plane $\textbf{M}_i$, with $\mathbf{T}_{\text{reflect}_i}$ describing the transformation induced by the mirror reflection, which not only enables multi-view geometry reconstruction using reflected appearances but also provides strong geometric priors for estimating virtual camera poses.

To this end, we formulate the task from a single-view 3D reconstruction problem to a multi-view 3D reconstruction problem, which can be effectively addressed using modern 3D reconstruction models, such as \reconmodel, which is designed to handle 3D reconstruction in a general setting. 
By leveraging our virtual view design, we adapt \reconmodel to perform 3D reconstruction under the single-view mirror-assisted setup. 

We first present the theory underlying the design of $\mathbf{T}_{\text{reflect}_i}$ in \Cref{sec:virtual-camera-design}, which simulates the physical imaging process of a virtual camera, ensuring that pixel-domain operations are equivalent to real-world image formation.
We then describe the full \ourmethod pipeline in \Cref{sec:pipeline} and introduce a symmetric-aware loss for post-optimisation in \Cref{sec:symmetric-aware-loss}, which leverages reflection constraints to further refine the estimated poses.

\subsection{Virtual View Imaging Principle}
\label{sec:virtual-camera-design}

\begin{figure}
    \centering
    \includegraphics[width=0.8\linewidth]{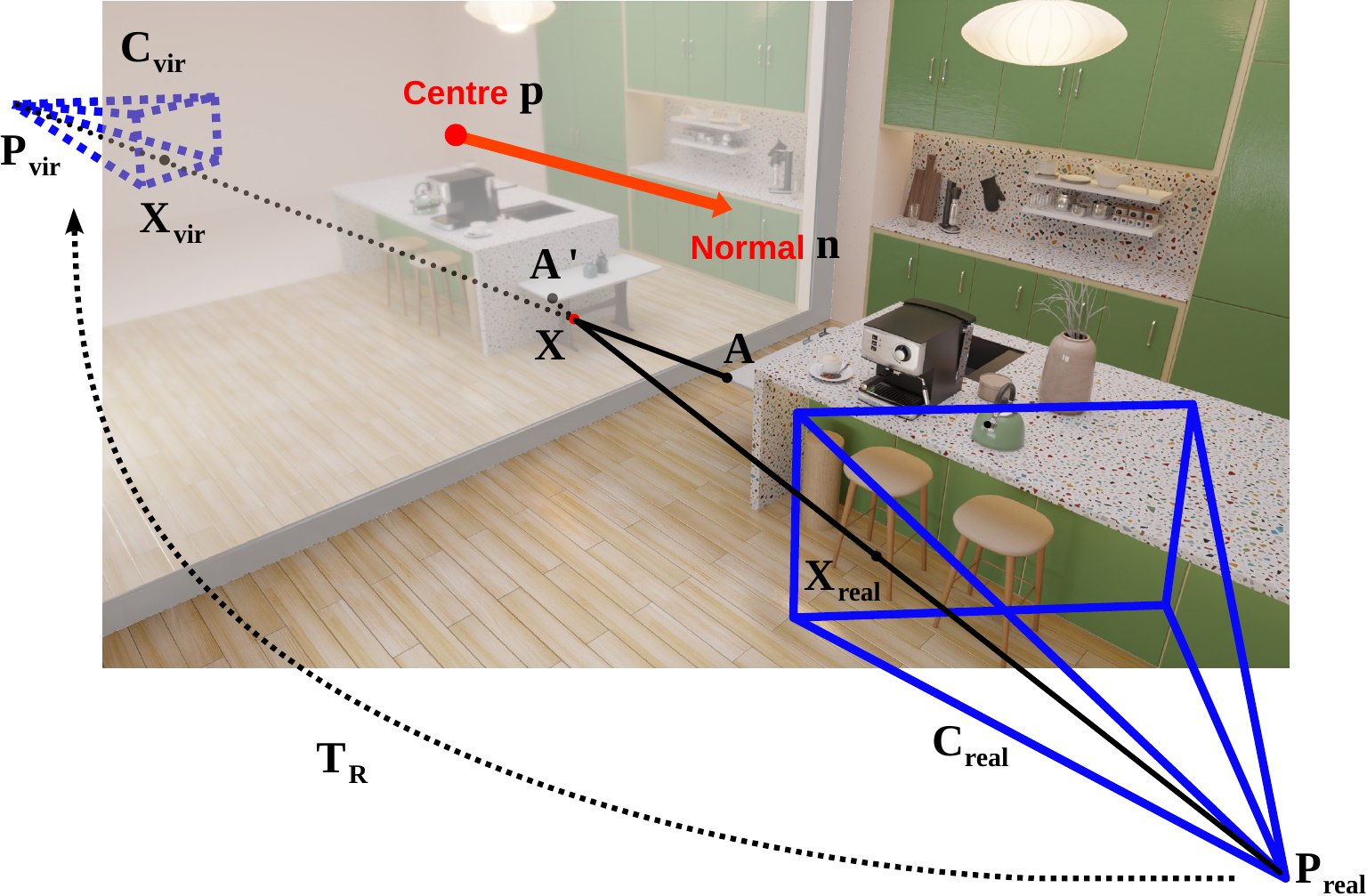}
    \caption{Physical imaging process of a scene containing a mirror. The reflection plane is shown as semi-transparent to reveal the virtual camera. 
    }
    \vspace{-0.6cm}
    \label{fig:theory}
\end{figure}

We first introduce the theory of simulating the virtual view imaging to obtain a multi-view setup, leveraging a mirror reflection to simplify 3D reconstruction. 
This approach allows us to directly flip the view in the pixel domain to create a valid virtual view, effectively transforming a single image containing mirrors into a multi-view input. 
The resulting view pairs can then be processed by advanced feed-forward 3D reconstruction models.

The physical imaging process is illustrated in \cref{fig:theory}. 
We define the reflection matrix $\textbf{T}_{\text{reflect}}$, which transforms a real camera $\textbf{C}_{\text{real}}$ (solid line) into its symmetric valid virtual camera $\mathbf{C}_{\text{vir}}$ (dashed line) across a mirror plane $\textbf{M}$. 
The reflection matrix is extended from the Householder matrix \cite{golub2012matrix}:
\begin{align} 
\label{equ:reflection}
\textbf{T}_{\text{reflect}} &= 
\textbf{diag}(-1,1,1,1) \cdot
\begin{bmatrix}
\mathbf{I} - 2\mathbf{n}\mathbf{n}^\top & 2( \mathbf{n}^\top \mathbf{p} ) \mathbf{n} \\
\mathbf{0}^\top & 1
\end{bmatrix}; \\
& = \textbf{diag}(-1,1,1,1) \cdot
\begin{bmatrix}
\mathbf{R}_{\text{reflect}} & \mathbf{t}_{\text{reflect}} \\
\mathbf{0}^\top & 1
\end{bmatrix},
\end{align}
where $\mathbf{n}, \mathbf{p}$ are the normal and the point on the mirror plane (centre in \cref{fig:theory}), respectively. 
$\textbf{diag}(-1,1,1,1)$ flips the $x$-axis of the coordinate system. 
Using this matrix, the virtual camera $\textbf{C}_{\text{vir}}$ is derived as:
\begin{equation} 
\label{equ:relation}
\textbf{C}_{\text{vir}} = 
\textbf{T}_{\text{reflect}} \cdot [\textbf{R}_{\text{real}} | \textbf{t}_{\text{real}}].
\end{equation}
For a 3D point $\mathbf{A} \in \mathbb{R}^{3\times1}$ visible to both cameras,
the real camera observes both the point $\mathbf{A}$ and its mirror reflection $\mathbf{A}'$, while the virtual camera observes only $\mathbf{A}$. 
The light rays $\mathbf{P}_{\text{real}}\mathbf{A}'$ and $\mathbf{P}_{\text{vir}}\mathbf{A}$ intersect at the same point on the mirror plane.
Let $\mathbf{X} \in \mathbb{R}^{3\times1}$ denote this intersection.
The projection of $\mathbf{A}$ onto the image plane of $\mathbf{C}_{\text{real}}$ is therefore equivalent to projecting $\mathbf{X}$ onto $\mathbf{C}_{\text{real}}$, given by:
\begin{align} 
\mathbf{X}_{\text{real}} = 
\begin{bmatrix}
u_{\text{real}}  \\
v_{\text{real}}  \\
1
\end{bmatrix} & = 
\mathbf{K} \cdot
[\textbf{R}_{\text{real}} | \textbf{t}_{\text{real}}] \cdot \mathbf{X}, 
\end{align}
where $\mathbf{K}$ is the intrinsic parameter of the real camera. 
Since the real and virtual cameras share the intrinsic parameter $\mathbf{K}$, the projection of $\textbf{A}$ on the virtual camera is:
\begin{align} 
\mathbf{X}_{\text{vir}} = 
\begin{bmatrix}
u_{\text{vir}}  \\
v_{\text{vir}}  \\
1
\end{bmatrix} & = \mathbf{K} \cdot \mathbf{T}_{\text{reflect}} \cdot 
[\textbf{R}_{\text{real}} | \textbf{t}_{\text{real}}] \cdot \mathbf{X}.  
\end{align}
Since the real and virtual cameras are symmetric with respect to the mirror plane, and $\textbf{diag}(-1,1,1,1)$ flips the $x$-axis, thus:
\begin{align}
    u_{\text{real}} + u_{\text{vir}} & = W;  \\
    v_{\text{real}} & = v_{\text{vir}},  
\end{align}
where $W$ is the width of the camera pixel plane. 

By flipping the real view horizontally, we ensure that the projection on the virtual camera is a flipped view of the real camera, maintaining the stereo setup needed for 3D reconstruction, enabling us to further leverage the reflection information as an auxiliary view.

\subsection{\ourmethod Pipeline}
\label{sec:pipeline}

\begin{figure*}
    \centering
    \includegraphics[width=1\linewidth]{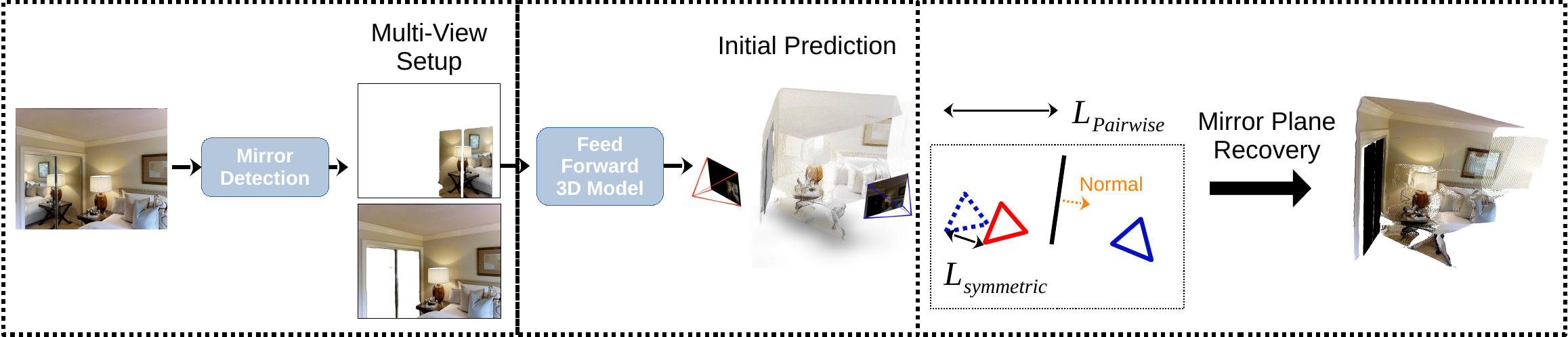}
    \caption{Overview of the proposed \ourmethod pipeline. \ourmethod reconstructs 3D scenes from a single-view image by leveraging mirror reflections. A reflection transformation is designed to ensure that flipping the real view in the pixel domain, simulating a virtual camera imaging, enables seamless integration with modern feed-forward models. Following the initial prediction, the reflection transformation is used as a geometric constraint to refine pose optimization. }
    \label{fig:method}
    \vspace{-0.5cm}
\end{figure*}

In this section, we introduce the pipeline of \ourmethod, which is illustrated in \cref{fig:method}. 

\boldstart{Mirror Detection and Multi-View Setup.} 
We first apply mirror detection to identify the reflection regions $\mathcal{F}$ in the image, using Detect Any Mirror (DAM) \cite{xing2023DAM}.
The detected regions are then flipped horizontally to simulate the imaging process of the virtual camera.
This allows the single input image $I$ to be decomposed into a real view and its corresponding reflected views.

\boldstart{Initial Prediction.}
We adopt \reconmodel as the backbone to generate the initial point cloud from the virtual-real pair, which processes one image pair at a time.
It first constructs a pairwise graph $\mathcal{E} = \{ e_i = (R, {V_i}) \mid i = 1, 2, \dots \}$, where $R$ denotes the real view and ${V_i}$ the $i$-th virtual view, and then predicts the corresponding point cloud set $\mathcal{S} = \{ S_{e} \mid e \in \mathcal{E} \}$.

\boldstart{Post-optimisation.}
We follow \reconmodel to perform the post-optimisation. 
To align point clouds into a global unified pointmap $\mathcal{U} = \{ U_i  \mid i = \{ R, V_1,V_2, ... \}$, \reconmodel adopts a pair-wise loss. 
For an input image pair $e = (R, V_j) \in \mathcal{E}$, \reconmodel first predicts the pointmaps $S_e = (S_{R; (R, V_j)}, S_{\text{vir}_j; (R, V_j)})$ and the confidence map $O_e = (O_{R; (R, V_j)}, O_{\text{vir}_j; (R, V_j)})$. 
\reconmodel introduces a pair-wise pose $P_e \in \mathbb{SE}(3)$ and a scaling factor $\sigma_e \in \mathbb{R}$ for each pair $e$ to rotate all pairwise predictions to align them to a shared global coordinate system $X_{\text{vir}}$. 
The alignment loss is formulated as:
\begin{equation}
    \mathcal{L}_{\text{pairwise}} = \sum_{e \in \mathcal{E}} \sum_{v \in e} \sum_{j=1}^{HW} O^j_{v} || U^{j}_v - \sigma_e P_e S^j_{v;e} ||.
\end{equation}

We extend the post-optimisation process with a \textbf{symmetric-aware loss} $\mathcal{L}_{\text{sym}}$ that constrains pose estimation and incorporates the geometric prior from our virtual camera design.
The final loss term $\mathcal{L}$ adopted for the post-optimization is
\begin{align}
    \mathcal{L} = \mathcal{L}_{\text{pairwise}} + \mathcal{L}_{\text{sym}}.
\end{align}

We recover the mirror plane from the result point cloud by estimating its normal and a point on the plane from the result point cloud. 
The details of the mirror plane recovery are covered in the supplementary. 


\subsection{Symmetric-Aware Loss}
\label{sec:symmetric-aware-loss}
For each pair $e$, the estimated camera poses $(\mathbf{C}_{\text{real}}, \mathbf{C}_{\text{vir}_j})$ are constrained to be symmetric with respect to a mirror plane $\textbf{M}_e$, estimated from the real view's point cloud $U_{\text{real}}$ using the mirror mask $M_{\text{real}}$.
Specifically, we estimate the normal $\mathbf{n}_e$ by performing Principal Component Analysis (PCA) of the mirror part of the point cloud $U_{\text{real}} \cdot M_{\text{real}}$.
We obtain the point $\mathbf{p}_e$ on the plane by computing the centroid of the masked point cloud $U_{\text{real}} \cdot M_{\text{real}}$.   

\begin{equation}
\mathbf{n}_e = \text{PCA}(U_{\text{real}} \cdot M_{\text{real}}); \quad \mathbf{p}_e = \overline{U_{\text{real}} \cdot M_{\text{real}}}.
\end{equation}

The symmetry constraint enforces that the virtual camera $\mathbf{C}_{\text{vir}_j}$ and the real camera $\mathbf{C}_{\text{real}}$ satisfy the relation in \cref{equ:relation}, as defined by the reflection transformation $\mathbf{T}_{\text{reflect}}$ in \cref{equ:reflection}.

To enforce this constraint, we formulate an optimization problem that minimizes the discrepancy between the transformed real camera pose $\mathbf{C}'_{\text{real}} = \mathbf{T}_{\text{reflect}} \cdot \mathbf{C}_{\text{real}}$ and the estimated virtual camera pose $\mathbf{C}_{\text{vir}_j}$ for all pairs in $\mathcal{E} = \{(R, V_j) \mid j=1,2, \dots \}$. The optimisation problem is defined as:
\begin{align}
\min_{\mathbf{C}_{\text{real}}, \mathbf{C}_{\text{vir}}} \quad & \sum_{j = 1 }^N \sum_{ (R, V_j) \in \mathcal{E}} || \mathbf{C}'_{\text{real}} \ominus \mathbf{C}_{\text{vir}_j} || \\
\text{s.t.} \quad & \mathbf{C}'_{\text{real}} = \mathbf{T}_{\text{reflect}} \cdot \mathbf{C}_{\text{real}}, \quad \forall (R, V_j) \in \mathcal{E},
\end{align}
where $\ominus$ denotes the difference between the transformed and estimated virtual camera poses.



We then compute the symmetric-aware loss by decomposing it into two components: the rotational difference and the translational difference:
\begin{align}
    \mathcal{L}_{\text{sym}} & =  \mathcal{L}_{\text{rot}} + \mathcal{L}_{\text{trans}}. 
\end{align}
The rotation term penalizes the angular difference between the rotation of $\mathbf{C}_{\text{vir}_j}$ and $\mathbf{C}'_{\text{real}}$, measured using quaternions:
\begin{equation}
    \mathcal{L}_{\text{rot}} =  \sum_{(D,V_j) \in \mathcal{E}} 1 - \mathbf{q'}_{\text{real}}^\top  \mathbf{q}_{\text{vir}_j};  \qquad ||\mathbf{q}||=1,
\end{equation}
where $\mathbf{q}'_{\text{real}}$ and $\mathbf{q}_{\text{vir}_j}$ are the unit quaternions corresponding to the rotation matrices of $\mathbf{C}_{\text{vir}_j}$ and $\mathbf{C}_{\text{real}}'$, respectively.


The translation term penalizes the Euclidean distance between the translation vectors:
\begin{align}
    \mathcal{L}_{\text{trans}} & = \sum_{(D,V_j) \in \mathcal{E}} || \mathbf{t}_{\text{real}}' - \mathbf{t}_{\text{vir}_j} ||_2^2.
\end{align}
This symmetric-aware formulation enforces consistency between the camera poses of the real and virtual views, guiding the optimisation toward a geometrically plausible configuration.





%% file: sec/4_dataset.tex
\section{Dataset}

\begin{figure}
    \centering
    \includegraphics[width=1\linewidth]{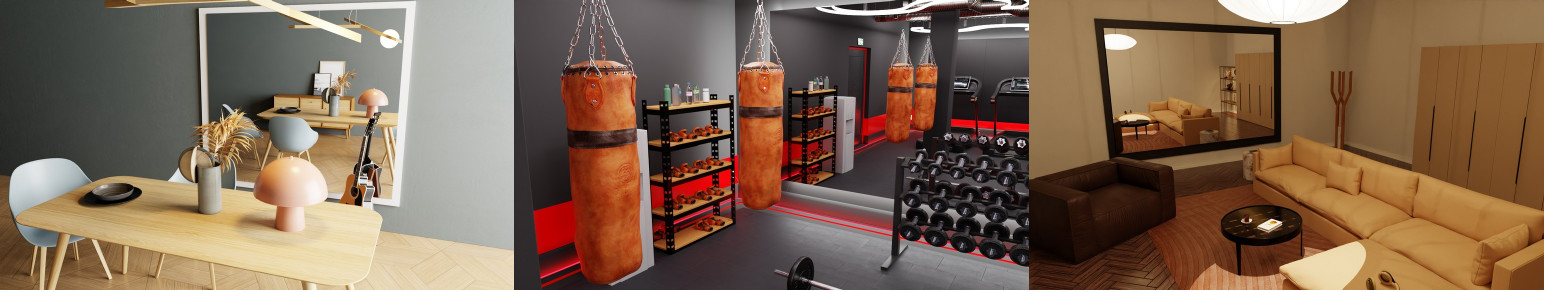}
    \caption{Thumbnails of the dataset, where each image represents a fully customizable Blender scene. }
    \label{fig:dataset}
    \vspace{-0.5cm}
\end{figure}

We construct a reflection-aided single-view reconstruction dataset of \synscenenum synthetic scenes sourced from Blender Demo \cite{blenderdemo}, BlenderKit \cite{blenderkit}, and CGTrader \cite{cgtrader}, manually modelled in Blender \cite{blender}.
Each scene is remodelled by us by adding a mirror surface positioned in a plausible location.
For some scenes, we also manually model additional details to ensure consistent richness across the dataset, avoiding cases where certain scenes lack sufficient complexity or objects.
Example scenes are shown in \cref{fig:dataset}, with a full dataset overview provided in the Appendix.

To support reflection-based reconstruction, we provide a Blender toolkit running in Blender scripting that simulates the virtual camera imaging process described in \Cref{sec:virtual-camera-design}.
Our dataset provides fully customizable Blender files with ground-truth point clouds for both real and mirror-reflected regions, along with real and virtual camera poses, while the editable nature of these scenes facilitates dataset extension for larger and more diverse future benchmarks.

%% file: sec/4_experiments.tex
\section{Experiments}

\subsection{Experiment Setup}
\label{sec:exp-setup}

\boldstart{Backbone. } 
\ourmethod builds on \reconmodel \cite{dust3r_cvpr24}, a feed-forward architecture for stereo-based 3D reconstruction.
Given an image pair, a shared Vision Transformer (ViT) encoder extracts tokens, which are refined through cross-attentive transformer decoders and decoded into pointmaps assigning 3D coordinates to each pixel, along with confidence maps estimating prediction reliability.

\boldstart{Baselines. }
We compare \ourmethod with three representative feed-forward 3D reconstruction models: DUSt3R \cite{dust3r_cvpr24}, MASt3R \cite{mast3r_arxiv24}, and VGGT \cite{wang2025vggt}, and one monocular depth estimation model MoGe \cite{wang2024moge}. 
For DUSt3R and MASt3R, which do not accept a single input image, we duplicate the prompt image as input, while for VGGT, we directly use the prompt image.


\boldstart{Dataset. }
Since mirrors are often small and may not provide meaningful reflections, we focus on scenes where reflections are informative.
We collect 16 real scenes with significant mirror reflections from the Mirror3D \cite{mirror3d2021tan}, ScanNet \cite{dai2017scannet}, Matterport3D \cite{Matterport3D}, and NYUv2 \cite{Silberman2012nyuv2} datasets to evaluate our method.
In addition to real data, we also report results on our proposed synthetic dataset, which provides accurate ground truth for computing quantitative metrics and performing ablation studies.

\boldstart{Evaluation Metrics.}
We evaluate reconstruction quality using 4 metrics: \textbf{completeness}, \textbf{accuracy}, \textbf{F1 score}, and \textbf{chamfer distance}.
Accuracy and completeness measure the percentage of reconstruction-to-ground-truth and ground-truth-to-reconstruction distances below a 1 cm threshold, respectively. 
The F1 score is computed as the harmonic mean of accuracy and completeness.
Chamfer Distance measures the similarity between two point sets by computing the average nearest-neighbour distance from each point in one set to the other, ensuring both sets are close in 3D space.
The math behind the metrics is provided in the supplementary. 

\subsection{Results}



\label{sec:qualitative-results}

\begin{figure}
    \centering
    \includegraphics[width=0.9\linewidth]{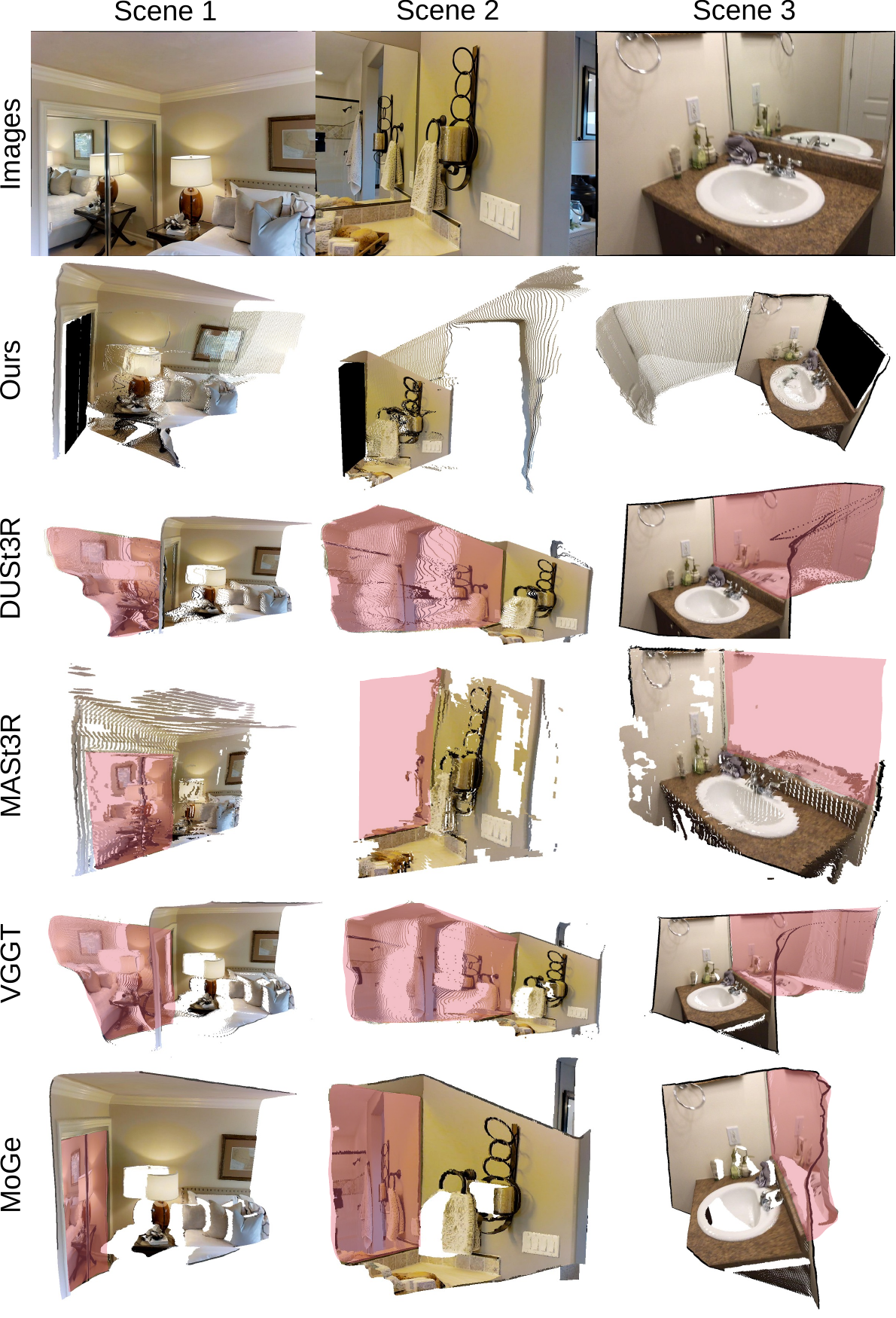}
    \caption{Qualitative results of \ourmethod and all the baselines running on the \textbf{real-world} data, where the predicted geometries corresponding to the mirror area are highlighted with \textcolor{purple}{light red}. }
    \label{fig:real-case}
    \vspace{-0.2cm}
\end{figure}

\begin{figure}
    \centering
    \includegraphics[width=0.9\linewidth]{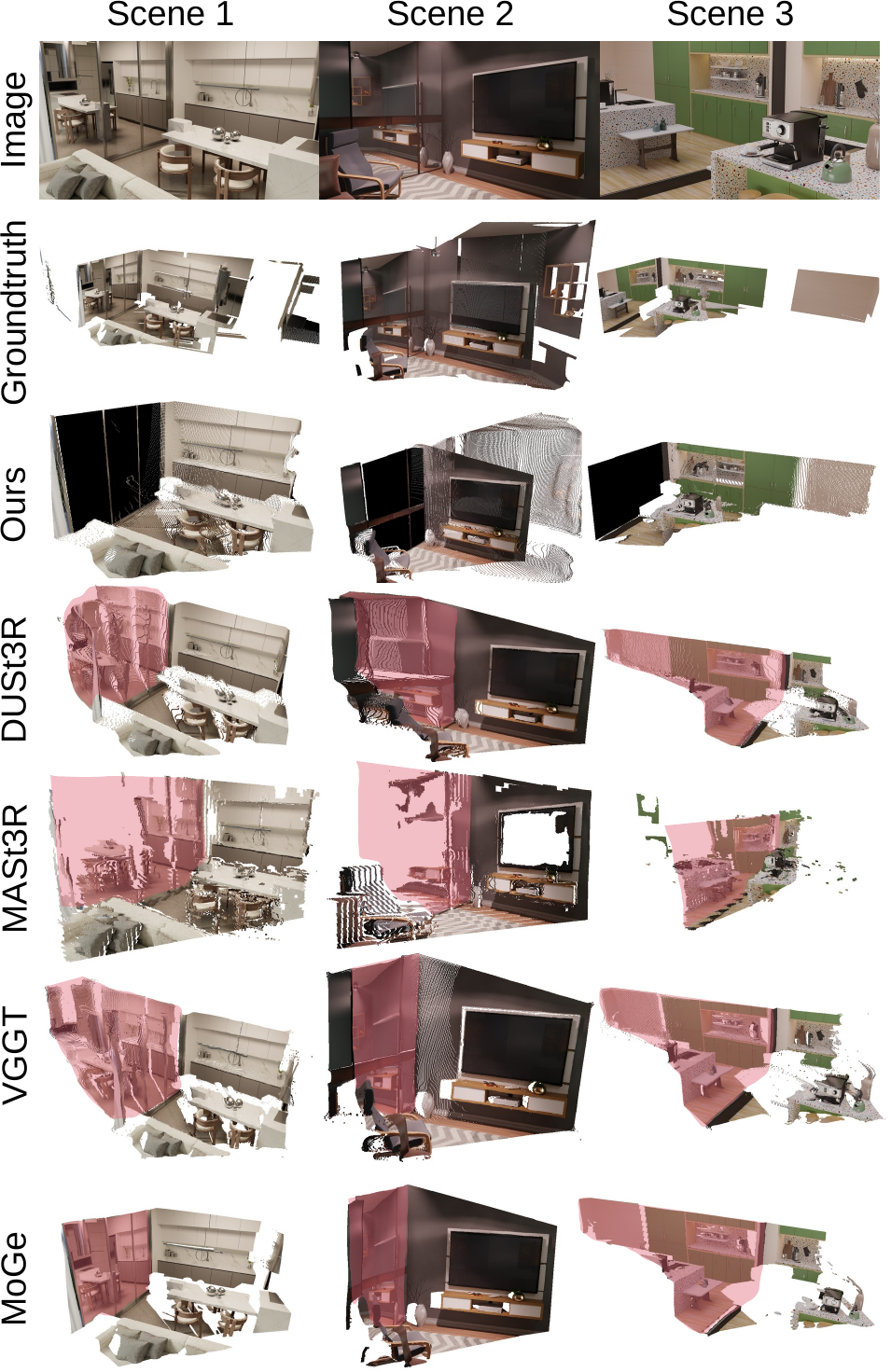}
    \caption{Qualitative results of \ourmethod and all the baselines running on the \textbf{synthetic} data, where the predicted geometries corresponding to the mirror area are highlighted with \textcolor{purple}{light red}. 
    }
    \label{fig:syn-case}
    \vspace{-0.2cm}
\end{figure}

\boldstart{Qualitative Results. }
We compare reconstructed 3D point clouds on both real and synthetic datasets in \cref{fig:real-case} and \cref{fig:syn-case}, where mirror regions are highlighted in \textcolor{purple}{light red} for all baselines.
In our method, the mirror region is correctly identified and represented as a black plane.
By reinterpreting reflections as virtual views captured by a virtual camera, \ourmethod introduces stereo information that significantly improves reconstruction coverage and completeness.
In contrast, \reconmodel, MASt3R, and VGGT all fail on both real and synthetic scenes: \reconmodel and VGGT misinterpret the mirror as false geometry (with the exception of VGGT on synthetic Scene 2), while MASt3R collapses to flat, degenerate geometry in all cases.
MoGe performs better at handling mirror regions compared to these baselines, yet it still produces erroneous geometry (e.g., real Scene 3 and synthetic Scene 3).

Moreover, the inability of the baselines to exploit stereo information prevents them from recovering occluded regions.
In contrast, our stereo formulation enables more complete reconstruction of hidden areas, for example, the back of the lamp (\cref{fig:real-case}, Scene 1), the rear surface of the towel (\cref{fig:real-case}, Scene 2), and the basin interior (\cref{fig:real-case}, Scene 3), all of which are effectively recovered by \ourmethod but missed by other methods.

\begin{table}[t]
\footnotesize
\centering
\begin{tabular}{lcccc}
\toprule
Method   & Comp. \% $\uparrow$ & Accuracy \% $\uparrow$ & F1 $\uparrow$ & Chamfer $\downarrow$ \\
\hline
Dust3r & 68.20\%           & 71.84\%               & 69.82\%  & 0.1021    \\
Mast3r & 37.80\%           & 62.62\%               & 45.42\%   & 0.1189   \\
VGGT   & 76.70\%           & 78.98\%               & 77.73\%    & 0.0906  \\
MoGe   & 79.50\%           & 88.51\%               & 83.69\%   & 0.0641 \\
\textbf{\ourmethod}   & \textbf{89.37}\%           & \textbf{96.64}\%               & \textbf{92.81}\%   & \textbf{0.0261}   \\
\bottomrule
\end{tabular}
\caption{Quantitative comparison of point cloud coverage between \ourmethod and all baselines on the \textbf{synthetic} dataset. Comp. and Chamfer denote completeness and chamfer distance, respectively. }
\vspace{-0.5cm}
\label{tab:baselines}
\end{table}




\boldstart{Quantitative Results. } 
We report quantitative comparisons against baseline methods in \cref{tab:baselines}, computing the mean across all the synthetic scenes for each metric.

\ourmethod consistently outperforms all baselines across all metrics.
It achieves a significant gain in completeness, exceeding the second-best baseline, MoGe (79.50\%), by roughly 10 percentage points.
Notably, \ourmethod simultaneously achieves high completeness, high accuracy, and low chamfer distance, demonstrating its ability to recover a larger portion of the scene while preserving geometric fidelity.
In contrast, MASt3R performs poorly due to its limited capability in sparse-view scenarios, and both DUSt3R and VGGT are adversely affected by the ambiguity introduced by mirrors, often misinterpreting reflections as real geometry.

These results demonstrate the effectiveness of reformulating the single-view input containing mirrors as a stereo setup: by leveraging the mirrored information, \ourmethod achieves denser and more accurate reconstructions. 
Reflected regions with overlapping content are reconstructed with higher confidence, while even non-overlapping or occluded areas, such as regions behind the camera or outside the field of view, benefit indirectly through improved global consistency and scene completion.


\begin{figure*}
    \centering
    \includegraphics[width=0.95\linewidth]{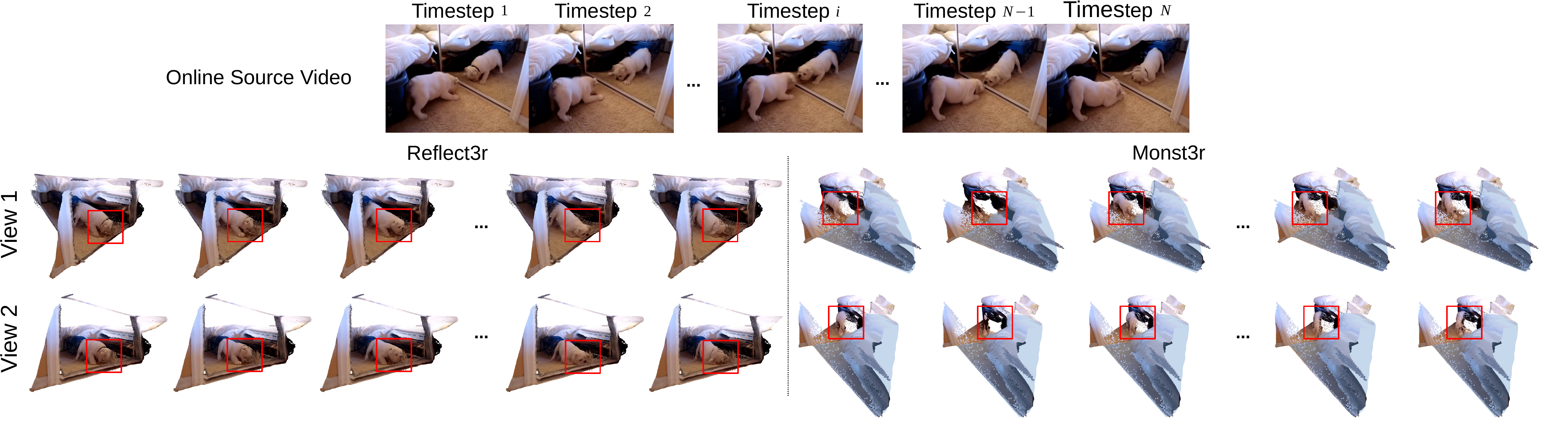}
    \caption{Dynamic reflection-aided 3D reconstruction. Comparison of single-view dynamic reconstruction results between \ourmethod and MonST3R on an online YouTube video. }
    \label{fig:dynamic}
    \vspace{-0.5cm}
\end{figure*}

\subsection{Ablation Study on Symmetric-Aware Loss}


\begin{figure}
    \centering
    \includegraphics[width=0.6\linewidth]{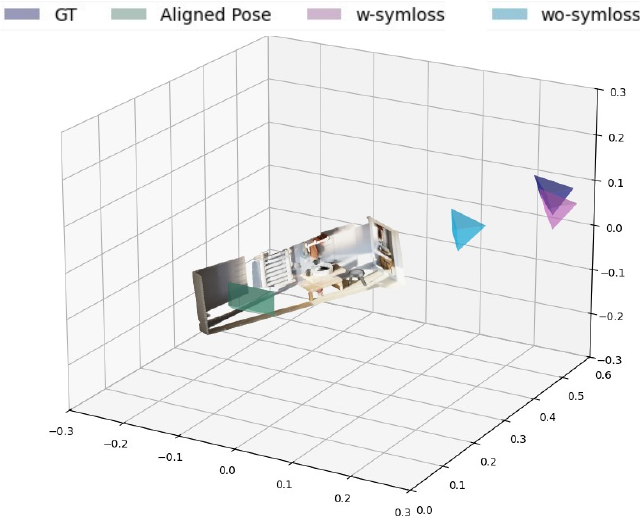}
    \caption{Comparison of pose estimation results with and without the symmetric-aware loss. \textcolor{violet}{Violet} indicates the ground-truth virtual pose, while \textcolor{pink}{pink} and \textcolor{cyan}{cyan} show the estimated poses with and without the symmetric-aware loss, respectively. 
    }
    \label{fig:ablation}
    \vspace{-0.5cm}
\end{figure}

We conduct an ablation study to evaluate the effectiveness of the symmetry-aware loss in refining camera pose optimisation.
As shown in \cref{tab:ablation}, we compare \ourmethod on the synthetic dataset with and without the symmetry-aware loss, using translation error ($T_{\text{err}}$) and rotation error ($R_{\text{err}}$) between predicted and ground-truth poses as evaluation metrics.
The results show a clear improvement in both the pose position (translation) and the pose orientation (rotation). 

\begin{table}[h]
\centering
\footnotesize
\begin{tabular}{lccll}
\hline
Setting & $T_{\text{err}}$ $\downarrow$   & $R_{\text{err}}$ ($ ^{\circ} $) $\downarrow$     \\
\hline
Without Sym-aware loss & 19.76\%	                  & 8.1624                   \\
With Sym-aware loss  & \textbf{13.58}\%                    & \textbf{5.8705}                   \\

\hline
\end{tabular}
\caption{Ablation study on the \textbf{synthetic} dataset evaluating the effectiveness of the symmetric-aware loss.
We report translation and rotation errors between the estimated and ground-truth poses, where translation error is the Euclidean distance between pose translations, and rotation error is the angular difference between pose rotations. 
}
\label{tab:ablation}
\vspace{-0.5cm}
\end{table}

In \Cref{fig:ablation}, we visualise a representative synthetic scene, comparing the optimised camera poses with and without the symmetry-aware loss.
The figure shows three sets of real–virtual camera pairs: ground-truth, estimates without the symmetry-aware loss, and estimates with the symmetry-aware loss.
For clarity, all real camera poses are aligned to a common reference pose (\textcolor{teal}{green}), making it easier to observe how the loss improves the pose estimation.

The translation error and the rotation error both decrease notably, showing the effectiveness of the symmetric-aware loss. 
With this loss, the predicted virtual camera poses (\textcolor{pink}{pink}) are more symmetric and better aligned with the ground-truth (\textcolor{violet}{violet}), while without it, the estimated reflected poses (\textcolor{cyan}{cyan}) deviate significantly.
This demonstrates that the symmetry-aware constraint helps enforce geometric consistency between the real and virtual views, leading to improved camera optimisation.


\subsection{Beyond the Static Scenes}





While our method is demonstrated on static scenes, the mirror-assisted reconstruction paradigm naturally extends to dynamic scene reconstruction.
Traditional dynamic 3D reconstruction requires synchronised multi-view images at each time step, which demands expensive hardware and precise camera synchronisation.

In contrast, mirrors offer a low-cost, scalable alternative, allowing multiple views to be captured simultaneously using a single camera.
Calibration is also simplified, as virtual cameras induced by mirrors inherently share the same intrinsic parameters as the real camera.


We demonstrate this concept using a dynamic version of \ourmethod, built upon MonST3R \cite{zhang2024monst3r}, a recent method designed for dynamic 3D reconstruction from a monocular video.
While MonST3R operates with only single-view information per frame, we augment it with virtual mirror views to provide multi-view cues at each time step.
The qualitative comparison between our dynamic \ourmethod and MonST3R is shown in \cref{fig:dynamic}, where we take an online YouTube video and perform the dynamic reconstruction for both methods.

Dynamic \ourmethod first detects the mirror in each frame and formulates the corresponding virtual view. 
Following MonST3R, we construct video graph pairs within a temporal window, and additionally introduce \textbf{spatial pairs} between the real and virtual views at each time step.
As virtual views tend to be more fragmented and have limited coverage, we construct temporal pairs exclusively between real views to ensure reliable motion estimation.
Specifically, for a video $\mathbf{V} = [\mathbf{I}^0, ..., \mathbf{I}^T]$, and a temporal window $w$, we define the view pair set at time step $t$ as, $\mathbf{W}^t = \{(\mathbf{I}^i_{\text{real}}, \mathbf{I}^i_{\text{vir}}), (\mathbf{I}_{\text{real}}^a, \mathbf{I}_{\text{real}}^b) | (a, b, i) \in [t, ..., t+w] \}$. 
Similar to \reconmodel, these pairs are then fed into MonST3R to obtain the initial point cloud prediction. 
In the post-optimisation process, we keep MonST3R’s temporal losses ($\mathcal{L}_{\text{smooth}}$, $\mathcal{L}_{\text{flow}}$) and incorporate our proposed symmetry-aware loss to improve spatial consistency between real and virtual views $(\mathbf{I}^i_{\text{real}}, \mathbf{I}^i_{\text{vir}})$ inside each frame. 

At each time step, our method produces dual-view reconstructions using only a single camera and a mirror, enabling temporally dense and geometrically complete 3D reconstruction with minimal calibration effort.

In contrast, relying only on monocular input, MonST3R fails to recover occluded regions such as the dog’s face (highlighted in \textcolor{red}{red} in \cref{fig:dynamic}), and misinterprets the mirror, producing incorrect geometry (highlighted in \textcolor{cyan}{light blue}).

%% file: sec/5_conclusion.tex
\section{Conclusion}


In this paper, we reformulate reflection-aided single-view reconstruction as stereo reconstruction by introducing a simple pixel-domain operation to leverage mirror reflections as auxiliary views.
This design enables a low-cost and easily deployable solution for both static and dynamic scene reconstructions. 
We propose \ourmethod, a pipeline that takes a single image as input, formulates virtual views via reflection, and reconstructs a more complete 3D point cloud via joint optimisation of geometry and camera poses.
Extensive experiments on real and synthetic data demonstrate significant improvements over existing baselines.
Additionally, we contribute a synthetic dataset of fully editable Blender scenes tailored for this setting to support future research.


%% file: sec/X_suppl.tex
\clearpage
\setcounter{page}{1}
\maketitlesupplementary

\section{Implementation Details}
\label{sec:recovery}
We explain more details about the mirror plane recovery in \cref{sec:exp-setup}. 
Since \reconmodel lacks direct correspondences in the mirror region, its predicted mirror plane is often inaccurate in position, though the normal direction is generally reliable.
Our objective is to recover both the plane’s normal and its position.
Our goal is to obtain the normal and the position of the plane. 
The normal is estimated using the method described in Eq. (11).
For plane positioning, we first extract edge points in the main view via image-space edge detection and back-project them into 3D.
RANSAC is then applied to remove outliers.
To improve robustness, we retain the top 10\% of edge points ranked by \reconmodel’s confidence scores and randomly select one as a reliable anchor point to finalize the plane position.

\section{More Details About the Synthetic Data}
\begin{figure}
    \centering
    \includegraphics[width=0.9\linewidth]{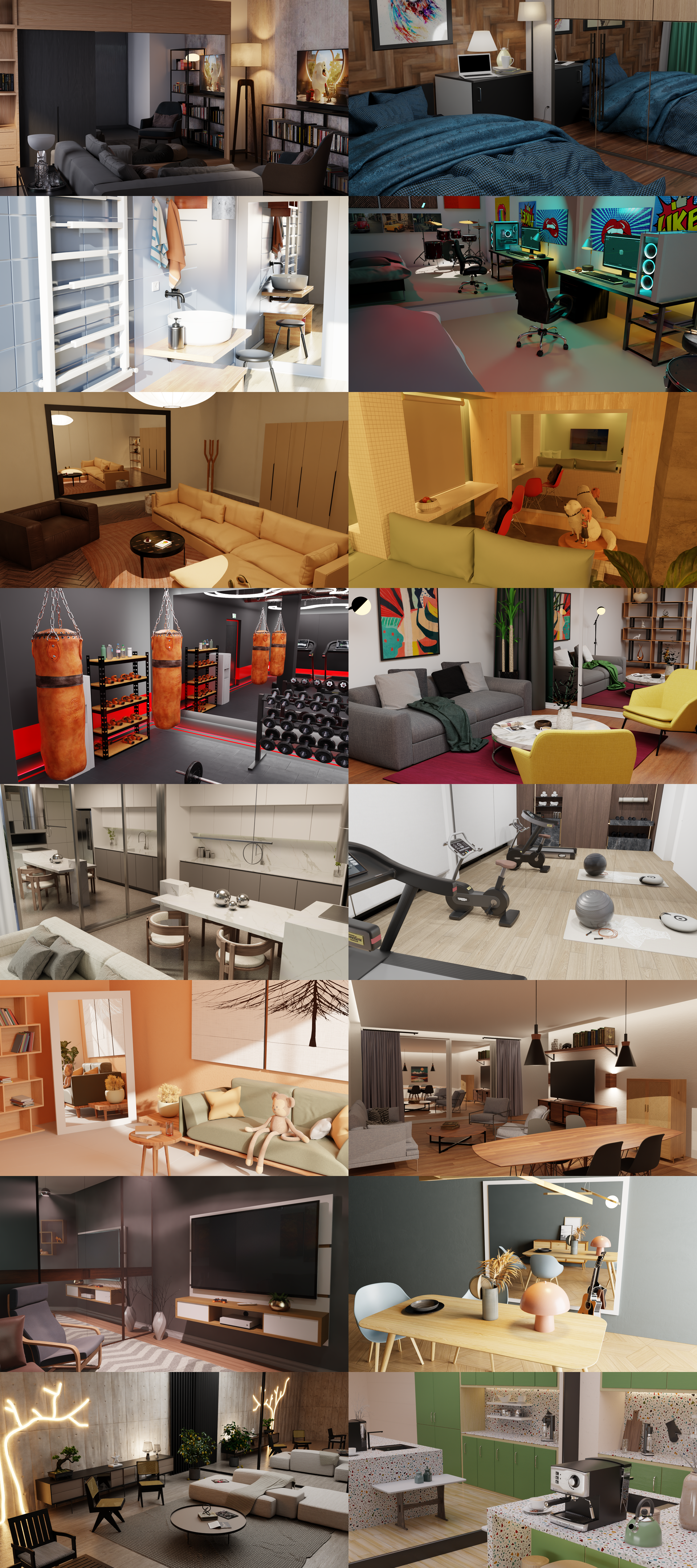}
    \caption{Thumbnails of the 16 fully editable Blender scenes included in our dataset.}
    \label{fig:whole-dataset}
    \vspace{-0.6cm}
\end{figure}

We provide the thumbnail of all 16 scenes included in our dataset in \cref{fig:whole-dataset}. 

\begin{figure}
    \centering
    \includegraphics[width=1\linewidth]{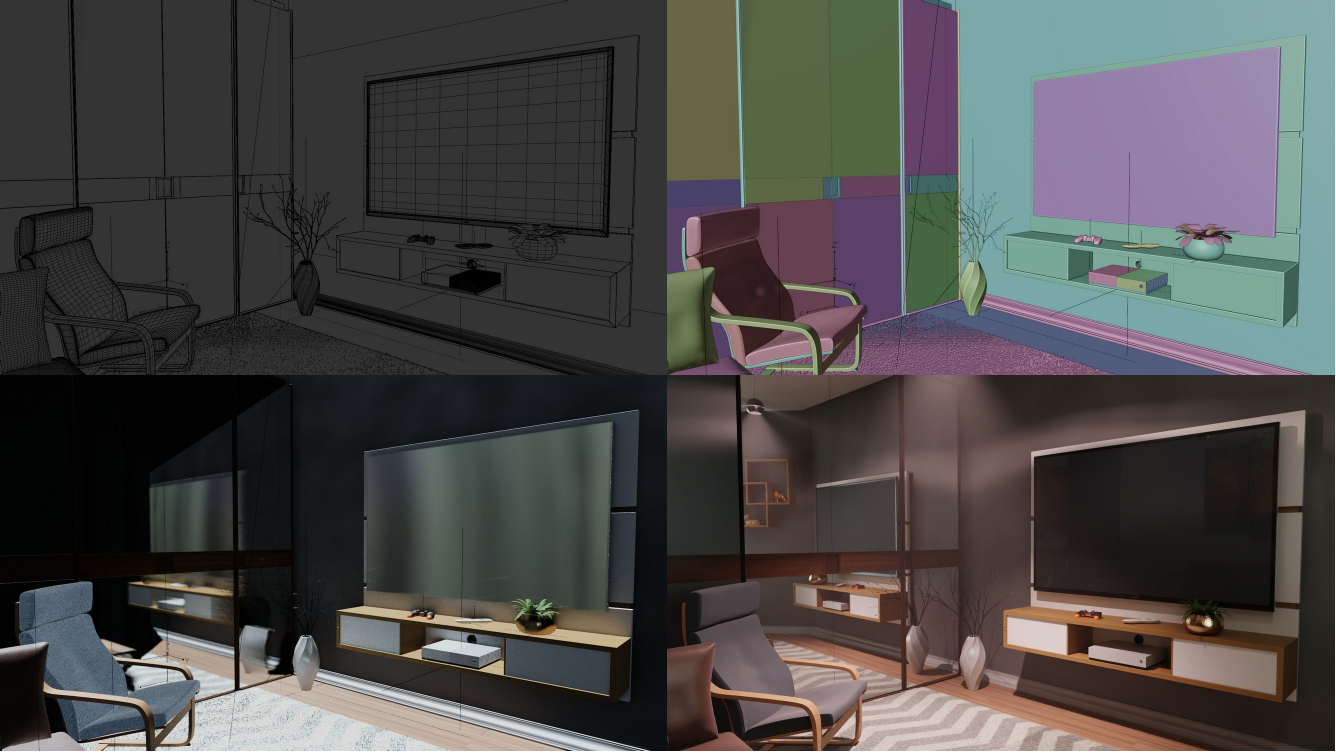}
    \caption{Details of the \textbf{synthetic} dataset’s Blender modeling. Top left: wireframe view; top right: solid view; bottom left: material view; bottom right: rendered view. The scene is fully customizable, allowing adjustments to the mirror’s position and properties, as well as the room setup and lighting, enabling easy extension of the dataset.}
    \label{fig:data-details}
    \vspace{-0.5cm}
\end{figure}

\Cref{fig:data-details} shows an example Blender scene in our dataset. 
Each scene in the dataset can be fully customized, including object shapes, room layouts, furniture placement, material properties, lighting conditions, etc.
We additionally insert and adjust mirror surfaces with controllable positions and reflectance properties to simulate realistic reflective setups.
This design not only ensures diverse and detailed scenes for training and evaluation but also provides a flexible foundation for extending the dataset in future work.

\section{More Details About the Experiments}

\subsection{Details of Evaluation Metrics}

We cover the math of the evaluation metrics in this section. 

Given the ground-truth point cloud $\mathcal{P} = \{ \mathbf{p}_i \}_{i=1}^{N_P}$ and the reconstructed point cloud $\mathcal{Q} = \{ \mathbf{q}_i \}_{i=1}^{N_Q} $,
Chamfer Distance is computed as
\begin{align}
    \frac{1}{2N_P}\sum_{\mathbf{p} \in \mathcal{P}} \min_{\mathbf{q} \in \mathcal{Q}} || \mathbf{p} - \mathbf{q} ||_2 + \frac{1}{2N_Q}\sum_{\mathbf{q} \in \mathcal{Q}} \min_{\mathbf{p} \in \mathcal{P}} || \mathbf{p} - \mathbf{q} ||_2
\end{align}

In our paper, we report the completeness (Comp.) and the accuracy (Accu.) in percentage with a threshold of 1 cm of the point cloud distance. 
Specifically, let the indicator function be $\mathbf{l}[\cdot]$,
\begin{align}
    \text{Comp.} &= \frac{1}{N_P} \sum_{\mathbf{p} \in \mathcal{P}} \mathbf{l}[ \min_{\mathbf{q} \in \mathcal{Q}} || \mathbf{p} - \mathbf{q} ||_2 < 1\text{cm} ]; \\
    \text{Accu.} &= \frac{1}{N_Q} \sum_{\mathbf{q} \in \mathcal{Q}} \mathbf{l}[ \min_{\mathbf{p} \in \mathcal{P}} || \mathbf{p} - \mathbf{q} ||_2 < 1\text{cm} ]; \\
    \text{F1} &= \frac{2 \cdot \text{Comp.} \cdot \text{Accu.}}{\text{Comp.} + \text{Accu.}}.
\end{align}

\begin{figure}
    \centering
    \includegraphics[width=0.8\linewidth]{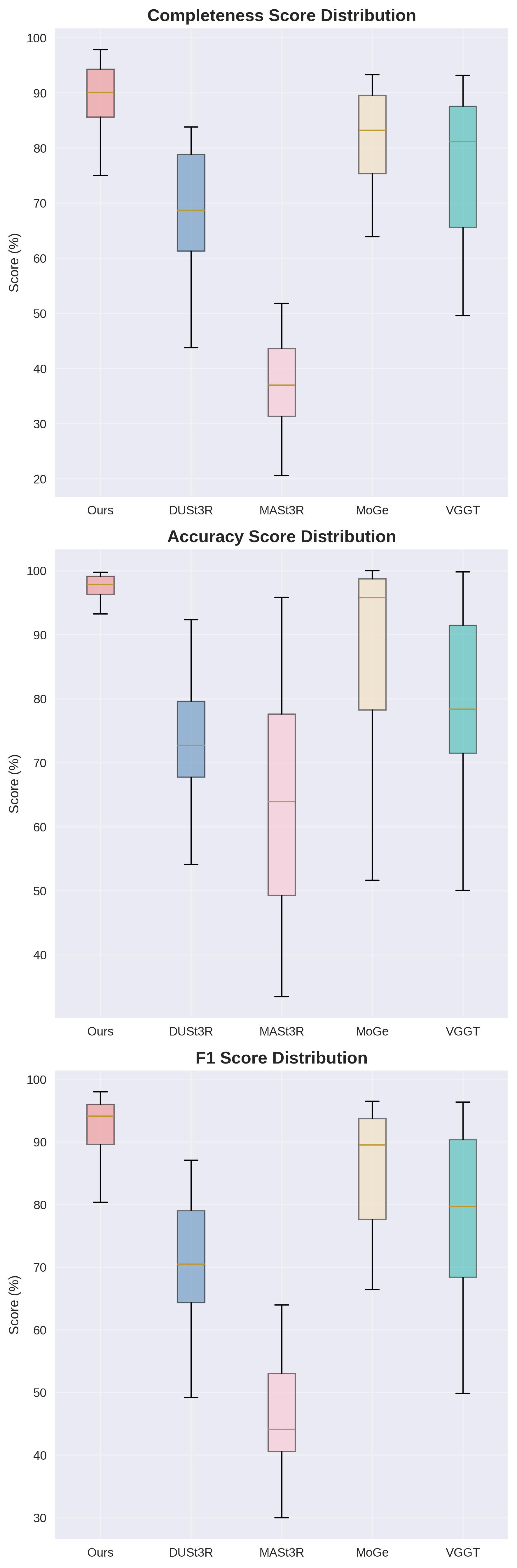}
    \caption{Boxplots of completeness, accuracy, and F1 score on the synthetic dataset.}
    \label{fig:exp-analysis}
\end{figure}

\subsection{More Statistics And Analysis}
To further assess robustness, we visualise the score distributions in \cref{fig:exp-analysis} using boxplots for completeness, accuracy, and F1 score on the synthetic dataset.
\ourmethod achieves the highest median scores across all metrics while also exhibiting the smallest interquartile range and whisker span, indicating low variance and strong stability across scenes.
MoGe attains similar median scores to \ourmethod, suggesting that its predicted point clouds are close to the ground-truth, but its results fluctuate significantly, reflecting high variance.
Its completeness is notably lower, as MoGe does not leverage the stereo information provided by mirrors and therefore fails to recover occluded regions, leaving parts of the scene uncovered.
DUSt3R and VGGT, unable to identify mirrors, generate false geometry in reflective regions and consequently show much higher variability.
MASt3R, which cannot handle single-view reconstruction, consistently predicts flat geometry and therefore yields uniformly low scores.